\documentclass{article}

\usepackage[final]{neurips_2023}

\usepackage[utf8]{inputenc} 
\usepackage[T1]{fontenc}    
\usepackage{hyperref}       
\usepackage{url}            
\usepackage{booktabs}       
\usepackage{amsfonts}       
\usepackage{nicefrac}       
\usepackage{microtype}      
\usepackage{xcolor}         
\usepackage{graphicx}
\usepackage{float}
\usepackage{natbib}
\usepackage{hyperref}
\usepackage{authblk}
\usepackage{graphicx}
\usepackage[mathscr]{eucal} 
\usepackage{amsbsy,amsmath,amssymb} 
\usepackage{empheq} 

\newcommand{\tensor}[1]{\boldsymbol{\mathscr{#1}}}   
\newcommand{\mat}[1]{\boldsymbol{{#1}}}   

\title{Can SAM recognize crops? Quantifying the zero-shot performance of a semantic segmentation foundation model on generating crop-type maps using satellite imagery for precision agriculture\thanks{Research was supported by the Agriculture and Food Research Initiative Competitive Grant no. 2020-69012-31914 from the USDA National Institute of Food and Agriculture and by the National Science Foundation CREST Center for Multidisciplinary Research Excellence in Cyber-Physical Infrastructure Systems (MECIS) grant no. 2112650.}}

\author[1]{Rutuja Gurav}
\author[1]{Het Patel}
\author[1]{Zhuocheng Shang}
\author[1]{Ahmed Eldawy}
\author[3]{Jia Chen}
\author[2]{Elia Scudiero}
\author[1]{Evangelos Papalexakis}

\affil[1]{Department of Computer Science and Engineering, UC Riverside, USA}
\affil[1]{\textit{\{rgura001, hpate061, zshan011, eldawy, epapalex\}@ucr.edu}}
\affil[2]{Environmental Sciences Department, UC Riverside, USA}
\affil[2]{\textit{elia.scudiero@ucr.edu}}
\affil[3]{Department of Electrical and Computer Engineering, UC Riverside, USA}
\affil[3]{\textit{jiac@ucr.edu}}

\begin{document}

\maketitle

\begin{abstract}
Climate change is increasingly disrupting worldwide agriculture, making global food production less reliable. To tackle the growing challenges in feeding the planet, cutting-edge management strategies, such as precision agriculture, empower farmers and decision-makers with rich and actionable information to increase the efficiency and sustainability of their farming practices. Crop-type maps are key information for decision-support tools but are challenging and costly to generate. We investigate the capabilities of Meta AI’s Segment Anything Model (SAM) for crop-map prediction task, acknowledging its recent successes at zero-shot image segmentation. However, SAM being limited to up-to 3 channel inputs and its zero-shot usage being class-agnostic in nature pose unique challenges in using it directly for crop-type mapping. We propose using clustering consensus metrics to assess SAM’s zero-shot performance in segmenting satellite imagery and producing crop-type maps.
Although direct crop-type mapping is challenging using SAM in zero-shot setting,
experiments reveal SAM’s potential for swiftly and accurately outlining fields in satellite images, serving as a foundation for subsequent crop classification.
This paper attempts to highlight a use-case of state-of-the-art image segmentation models like SAM for crop-type mapping and related specific needs of the agriculture industry, offering a potential avenue for automatic, efficient, and cost-effective data products for precision agriculture practices.
\end{abstract}

\section{Introduction}

Precision agriculture relies heavily on the accuracy of crop-type maps, as they serve as the foundation for informed decision-making in farming practices \cite{becker2023crop}. High-quality crop-type maps enable farmers to optimize resource allocation, monitor crop health, and maximize yields while minimizing environmental impacts. However, generating accurate crop-type maps is a resource-intensive and expensive endeavor, often requiring laborious manual annotation or sophisticated supervised deep learning models. Therefore, there is an ongoing effort at the confluence of precision agriculture and deep learning to develop efficient and reliable automated methods for crop-type map prediction using abundant remote sensing satellite imagery \cite{qadeer2021spatio}.

Meta AI's state-of-the-art Segment Anything Model (SAM) \cite{kirillov2023segany} has garnered significant attention for its remarkable performance in automatically segmenting various types of images, including natural scenes, medical images, and satellite images \cite{mazurowski2023segment, wang2023scaling, jing2023segment}. SAM, with its \textit{prompt-based interface} and \textit{automatic mask generator} \ref{bkgnd_sam}, has showcased impressive results even in zero-shot settings. Nevertheless, applying SAM to the challenging task of predicting crop-type maps presents unique challenges.

SAM is limited to images of up to 3 channels and was trained on an extensive dataset of RGB images. One of the primary difficulties lies in this inherent limitations of using only the RGB spectra of a rich, multi-spectral satellite imagery stack. Distinguishing between different crop types using only spectral information from RGB channels is challenging as crops often exhibit similar color characteristics, especially during early growth stages. Moreover, crop-type maps are traditionally produced using the temporal evolution of the normalized difference vegetation index (NDVI) over the whole growing season \cite{wei2023early, ghosh2021attention} and not just using an RGB snapshot of the crop fields at a single moment in time. Furthermore, SAM's class-agnostic nature complicates the direct application of its zero-shot automatic mask generator to generate crop-type maps as, unlike typical image segmentation models, it does not provide labels for pixels and instead outputs a set of boolean masks. 
This paper seeks to investigate these challenges by proposing the use of clustering consensus metrics to quantify SAM's zero-shot performance on the task. While direct crop-type map generation may be challenging, we envision leveraging SAM's strengths to produce fast and accurate shape maps outlining individual fields within a large agricultural area of interest in a satellite image. These shape maps, despite not directly representing crop types, can serve as a valuable foundation for subsequent crop type classification and map generation processes.

In this paper, we will present the methodology and experiments conducted to assess SAM's performance, highlighting the insights gained from using clustering consensus metrics. The rest of this paper is organized as follows - In Section \ref{sec:bkgnd}, we setup the preliminaries for a brief overview of crop-type mapping using remote sensing imagery and the Segment Anything model, followed by our experimental setup and analysis in Section \ref{sec:analysis}. Finally, we conclude with our findings and specify some future directions in Section \ref{sec:conc}.

\section{Background}
\label{sec:bkgnd}


\paragraph{Terminology:}
\begin{enumerate}
    \item \textbf{AOI:} Area of Interest; depending on the spatial resolution at which the satellite captures data, each pixel in the AOI represents physical land area (typically measured in meters$^2$).
    \item \textbf{NDVI:} Normalized difference vegetation index; it is a measure of "greenness" which quantifies vegetation by measuring the difference between near-infrared (which vegetation strongly reflects) and red light (which vegetation absorbs).
    
    \[NDVI_{Sentinel2} = \frac{(B8-B4)}{(B8+B4)} \]
    
    where $B8$: near-infrared band and $B4$: red band of the Sentinel-2 satellite measurements.

\end{enumerate}
\subsection{Crop Data  Layer (CDL): Data Product for high-quality crop-type maps}
\label{bkgnd_cdl}
The Cropland Data Layer (CDL) \cite{boryan2011monitoring}, hosted on \href{https://data.nal.usda.gov/dataset/cropscape-cropland-data-layer}{CropScape} \cite{han2012cropscape}, provides a raster, geo-referenced, crop-specific land cover map for the continental United States. CDL is an annual data product created at the end of the growing season by the US Department of Agriculture (USDA), and National Agricultural Statistics Service (NASS). It is produced at a 30 m resolution and provides pixel-level classification across several hundred crop-types grown in the US.

\subsection{Sentinel-2 Satellite Imagery}
\label{bkgnd_sentinel2}
The European Space Agency (ESA) provides an open release of the multi-spectral spatio-temporal earth observation data captured by their Sentinel-2 satellites \cite{drusch2012sentinel} at 10 m, 20 m and 60 m resolutions across visible, near infrared, and short wave infrared bands of the spectrum. In \cite{ghosh2021calcrop21}, the authors release the CalCrop21 dataset, which contains a portion of the Sentinel-2 data as well as the corresponding CDL (available \href{https://drive.google.com/drive/folders/1EnXXRHNoTyIbM-_5p-P9pH4zH3xyTqBp?usp=drive_link}{here}). The datasets consists of 367 \textit{tiles} (i.e. samples), each representing 1098 pixels x 1098 pixels AOIs spanning agricultural fields of Central Valley, California. Each sample represents the multi-spectral spatio-temporal stack of the AOI for the entire growing season of the year 2018. We will use this dataset for our analyses in this paper. See section \ref{analysis:dataset} for more details.

\subsection{Automatic mask generation with SAM}
\label{bkgnd_sam}
SAM is the latest amongst the so-called \textit{foundation models} for the crucial computer vision task of image segmentation (i.e. pixel classification). It is trained on the massive SA-1B dataset \cite{kirillov2023segany} consisting of 11 million RGB images and 1 billion masks resulting in its state-of-the-art performance in zero-shot setting on a variety of tasks. However, there are a few key ways in which SAM differs from traditional segmentation models that play a crucial role in its proposed use-case for our task -
\begin{enumerate}
    \item \textbf{Zero-shot inference:} SAM has been trained on a massive dataset and can be used in the so-called zero-shot setting i.e. without training or fine-tuning on a large, task-specific dataset.
    \item \textbf{Prompt-based interface:} SAM is designed to segment objects in an input image based on a set of \textbf{\textit{prompts}}. The prompts can be in form of \textit{points} and/or \textit{boxes} that a user can provide for a given image which can guide the model to isolate and segment objects in/around the prompted region in the input image.
    \item \textbf{Lack of class labels:} SAM is \textbf{\textit{class-agnostic}}. Its output is a set of boolean masks and it does not identify the objects it segments with any semantic class labels.
\end{enumerate}

In its \textit{segment everything} setting, lacking any user provided prompts indicating a region-of-interest within the image to be segmented, SAM's \textbf{Automatic Mask Generator (AMG)} returns a set of boolean masks (and associated metadata like predicted Intersection over Union (IoU) score)
given an input image by prompting the model with a grid of \textbf{\textit{uniformly distributed}} point prompts. There are a few tunable parameters available in the AMG that are relevant to our analysis - 
\begin{enumerate}
    \item \textbf{Points per side (PPS):} The number of points to be sampled along one side of the image. The total number of points is PPS$^{2}$. Higher PPS values ensure more unambiguous point prompts available per region in the input image at the cost of higher mask generation time.
    \item \textbf{Minimum Mask Region Area (MMRA):} Removes disconnected regions and holes in masks with area smaller than the MMRA value.
\end{enumerate}


\begin{figure}[]
  \centering
  \includegraphics[width=0.85\textwidth, trim={0 1.2cm 0 10.8cm}, clip]{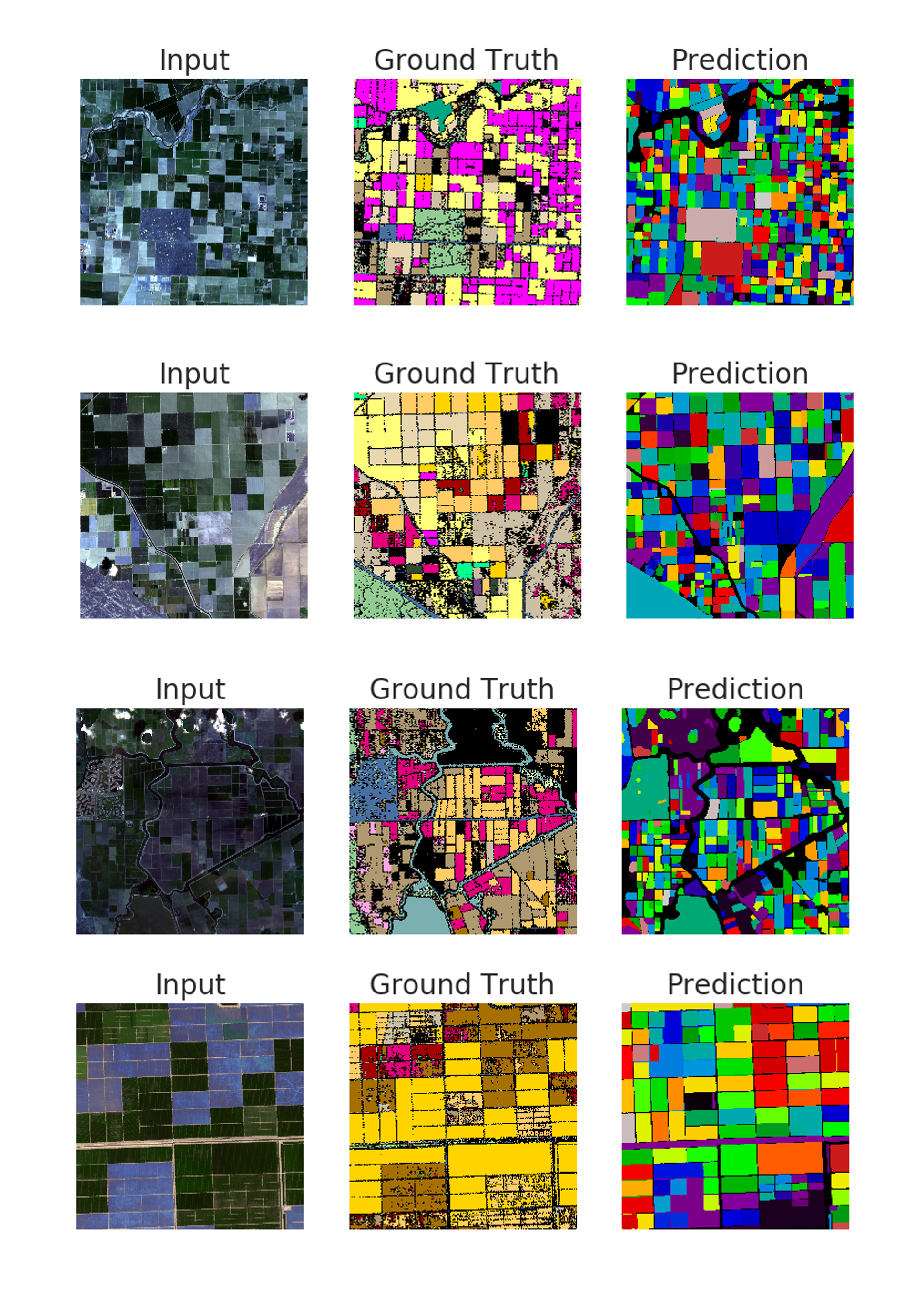}
  \caption{(left) Examples of input images (AOI = 1098 x 1098) created using the red-green-blue channels at the maximum NDVI timestep from the 4D multispectral spatiotemporal imagery stack from Sentinel-2 satellite, (middle) the ground-truth crop-type maps (CDL) depicting crop types and other related classes, (right) zero-shot predicted masks using SAM's automatic mask generator.}
  \label{fig:sample_pred}
\end{figure}

\section{Analysis}
\label{sec:analysis}
\subsection{Dataset}
\label{analysis:dataset}

A sample in the CalCrop21 dataset is a 4-D tensor $\tensor{X} \in \mathbb{R}^{T \times W \times H \times C}, \mat{Y} \in \mathbb{I}^{W \times H}$ where $W$ = image width, $H$ = image height, $T$ = no. of timesteps, $C$ = no. of spectral channels representing the multispectral spatio-temporal stack ($\tensor{X}$) that spans a 1098 x 1098 pixels AOI over 24 timesteps and across 10 spectral channels and the corresponding CDL ($\mat{Y}$). Whereas, SAM is limited to upto 3 channel inputs and was trained on the massive \textit{SA-1B} dataset of 11 million RGB images. Thus, we are limited to using the red, green and blue channels from our samples. Furthermore, out of the 24 timesteps available, we choose the timestep where the NDVI is maximum to produce a temporal snapshot of the crop fields when the crops are at peak "greenness". With this limited choices, we compute $\mat{X}_{RGB}$.
\[ \mat{X}_{RGB} = \tensor{X}(t_{max},:,:,[\textcolor{red}{i},\textcolor{green}{j},\textcolor{blue}{k}]) \] 
where $\mat{X}_{RGB} \in \mathbb{R}^{H \times W \times 3}$ such that $t_{max} \in T$ is the maximum NDVI timestep and $i,j,k \in C$ correspond to the \textcolor{red}{red}, \textcolor{green}{green}, \textcolor{blue}{blue} channels respectively. We use the ground-truth ($\mat{Y}$) as is. Thus, a sample representing the input image provided to SAM and the corresponding ground-truth CDL used to evaluate the prediction performance are ($\mat{X}_{RGB}, \mat{Y}$). See figure \ref{fig:sample_pred}; the left and middle plots show an example input image and the corresponding ground-truth CDL.

Note: The CalCrop21 dataset has 367 samples, however, after computing $\mat{X}_{RGB}$ we have deemed 20 of those tiles unusable due to the cloud cover present at the maximum NDVI timestep ($t_{max}$).

\subsection{Testing SAM's Automatic Mask Generator for crop-maps prediction}

Crop-map prediction requires us to assign a crop-type label to each pixel in the image. Traditionally, this translates to the popular multi-class semantic segmentation task that's well-studied in computer vision.
However, \textbf{SAM is class-agnostic}. As described in section \ref{bkgnd_sam}, when used in zero-shot, segment-everything, uniformly-prompted setting, \textbf{it produces a set of boolean masks and we cannot know the one-to-one mapping between them and the crop-type classes in the ground-truth CDL}. Therefore, we cannot compute the traditional evaluation metrics used in supervised learning settings viz. accuracy, dice coefficient, Intersection-Over-Union (IoU). To evaluate the quality of the predicted masks compared to the ground-truth mask, first, we have to post-process this collection of boolean masks into a single multi-class mask\footnote{For a pixel that belongs to more than one boolean mask (i.e. if a subset of masks overlap), we assign that pixel to the mask which has the highest predicted IoU score.}. 

Furthermore, there can be different no. of unique classes/labels in the ground-truth CDL compared to the predicted post-processed multi-class mask. Therefore, we have chosen to use \textbf{clustering consensus metrics} to quantify the agreement between the ground truth and predicted post-processed multi-class mask\footnote{The term "multi-class mask" is abused here to mean more than 2 classes to contrast with "binary mask" for which we have one-to-one mappings between ground-truth and predicted masks and can therefore calculate traditional segmentation metrics like dice coefficient and IoU.}. We flatten the ground-truth and predicted multi-class masks then treat them as two sets of clusterings of the pixels in the input image. A consensus metric would thus quantify agreement between the two sets, providing us an indirect measure of how closely SAM can predict the ground-truth CDL using just RGB images of the crop fields. We evaluate the multi-class mask that we derive from SAM's output on a variety of different clustering consensus metrics across varying $\sqrt{AOI}$ with respect to prompts density and minimum mask region area as a fraction of image length and image area respectively.
\paragraph{\textit{Observation:}} Preliminary testing suggests that \textbf{SAM can segment a semantically identical region (i.e. belonging to a single ground-truth class) as a single mask if that region remains spatially contiguous and occupies a relatively large fraction of the AOI in the input image} (See Appendix \ref{ap:examples_preds}). We'll term this type of samples easier for SAM to segment as expected\footnote{It is to be noted, of course, that we cannot establish a one-to-one mapping between the ground-truth classes and the labels in the predicted labels as SAM's predictions are entirely zero-shot and unsupervised.}. 

\begin{figure}[t!]
\begin{center}
  \includegraphics[width=0.85\textwidth]{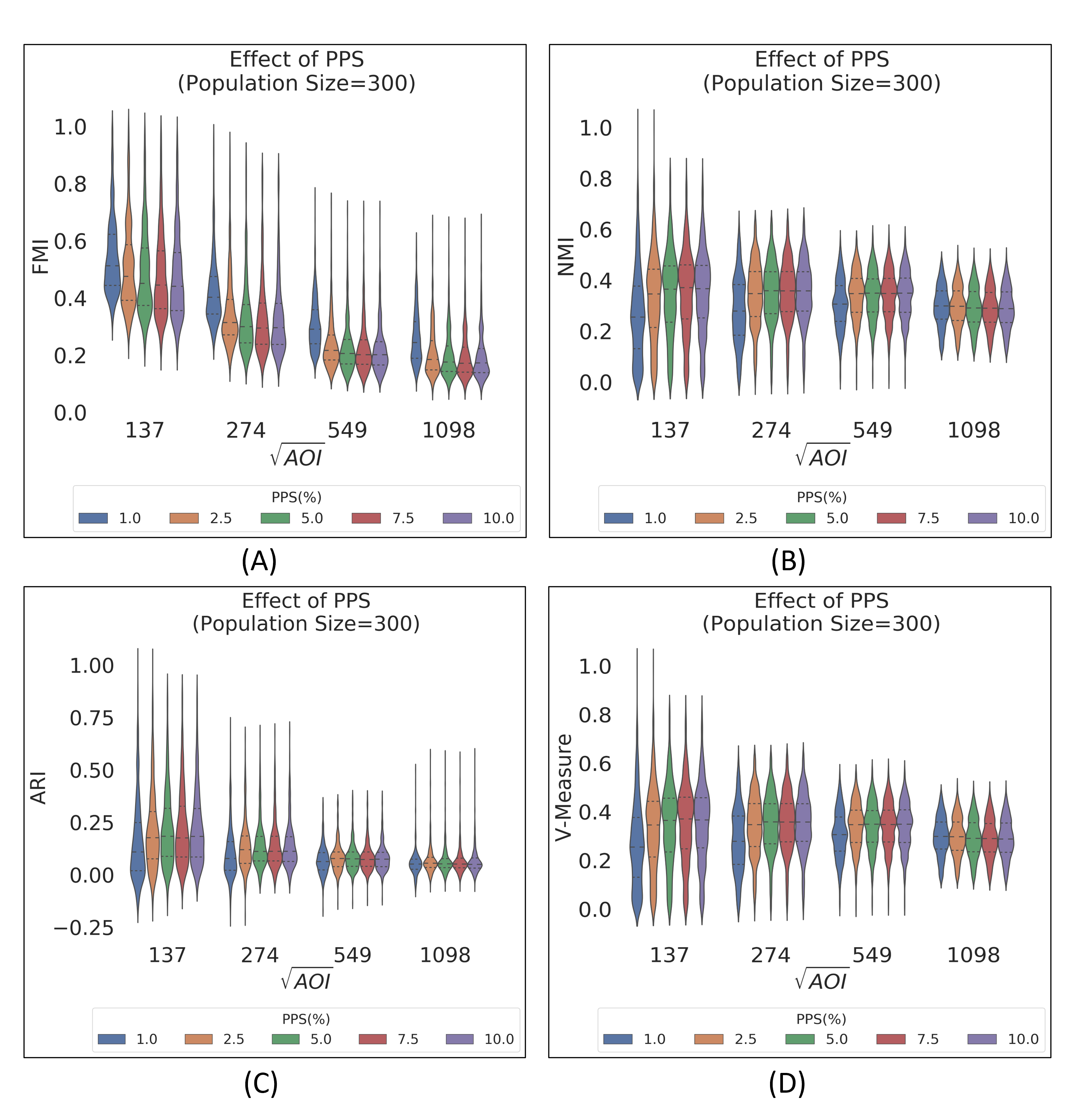} 
  \caption{Aggregate plots of clustering consensus between ground-truth CDL and multi-class mask produced by SAM's Automatic Mask Generator across 4 different metrics - \textbf{(A)} FMI: Fowlkes-Mallows Index; declines with increasing $\sqrt{AOI}$ as well as over increasing PPS\% for a given $\sqrt{AOI}$, \textbf{(B)} NMI: Normalized Mutual Information; unchanged with increasing $\sqrt{AOI}$, \textbf{(C)} ARI: Adjusted Rand Index; declines with increasing $\sqrt{AOI}$, \textbf{(D)} V-Measure: Correlated to NMI.}
  \end{center}
  \label{fig:perc_pps_aggs}
\end{figure}

The original tiles in the CalCrop21 dataset have $\sqrt{AOI} = 1098$ pixels and we have created sub-tiles with 2x, 4x and 8x smaller $\sqrt{AOI}$s with an overlapping sliding window over the original tiles to create easier samples as the $\sqrt{AOI}$ decreases. This results in samples with sub-tile dimensions (549 x 549), (274 x 274) and (137 x 137) respectively. We randomly select 300 samples from each set to perform our analysis. As shown in figure \ref{fig:perc_pps_aggs}, we observe some indicative trends across 4 different metrics. There are two types of trends to be noted - 1. overall trend across varying $\sqrt{AOI}$, 2. trend for a specific $\sqrt{AOI}$ across varying PPS\%.

\paragraph{Fowlkes-Mallows Index (FMI):}
FMI measures the geometric mean of precision and recall between the ground-truth and predicted clusters. 
FMI is sensitive to the number of true positive pairs and penalizes both false positives and false negatives. FMI ranges from 0 (random) to 1 (perfect consensus).
As shown in figure \ref{fig:perc_pps_aggs} (A), the relative decrease in mean FMI at increasing values of $\sqrt{AOI}$ suggests that the predicted clusters are becoming less accurate in terms of both false positives (pairs that are in the same predicted cluster but not in the same ground-truth cluster) and false negatives (pairs that are in the same ground-truth cluster but not in the same predicted cluster). The relative decrease in FMI for a given $\sqrt{AOI}$ at increasing values of PPS\% can be potentially explained by insufficient prompts density at the lower PPS\% leading to singular large mask that lowers the precision and/or recall.

\paragraph{Adjusted Rand Index (ARI):}
ARI quantifies the agreement between pairs of data points in terms of whether they are in the same or different clusters in both ground-truth and predicted clusterings while accounting for chance. ARI ranges from -1 (no agreement) to 1 (perfect agreement), with 0 indicating random agreement.
As shown in figure \ref{fig:perc_pps_aggs} (C), the relative decrease in mean ARI at increasing values of $\sqrt{AOI}$ indicates that the clustering is poor at capturing the overall structure of the data w.r.t. the ground-truth. The relative increase in mean ARI for a given $\sqrt{AOI}$ at increasing values of PPS\% suggests that the lowest prompt density value is not sufficient but the performance plateaus at the higher prompt densities suggesting no further gains can be made in terms of pairwise agreements between pixels in terms of cluster membership in ground-truth versus the prediction.
\paragraph{V-Measure:}
V-Measure is the harmonic mean of homogeneity and completeness, capturing both the quality of individual clusters and how well they cover the ground-truth classes.
Homogeneity measures whether each cluster contains only data points that are members of a single ground-truth class.
Completeness measures whether all data points that are members of a given ground-truth class are assigned to the same cluster.
As shown in figure \ref{fig:perc_pps_aggs} (D), the mean V-Measure remains the same across increasing values of $\sqrt{AOI}$ indicates that the predicted clusters are not becoming more homogeneous and complete as the $\sqrt{AOI}$ increases. The relative increase in mean V-Measure for a given $\sqrt{AOI}$ at increasing values of PPS\% suggests that the lowest prompt density value is not sufficient, but the performance plateaus at the higher prompt densities suggesting no further gains can be made in terms of homogeneity and completeness. 

\paragraph{Normalized Mutual Information (NMI):}
NMI measures the mutual information between the ground-truth and predicted clusters, normalized by entropy terms. It quantifies the amount of information shared between the two clusterings. NMI ranges from 0 (no mutual information) to 1 (perfect agreement). NMI is correlated to V-Measure, so as shown in figure \ref{fig:perc_pps_aggs}(B), we see similar trends at differing $\sqrt{AOI}$ and PPS\%.

\begin{figure}[!htp]
  \centering
  \includegraphics[width=0.9\textwidth]{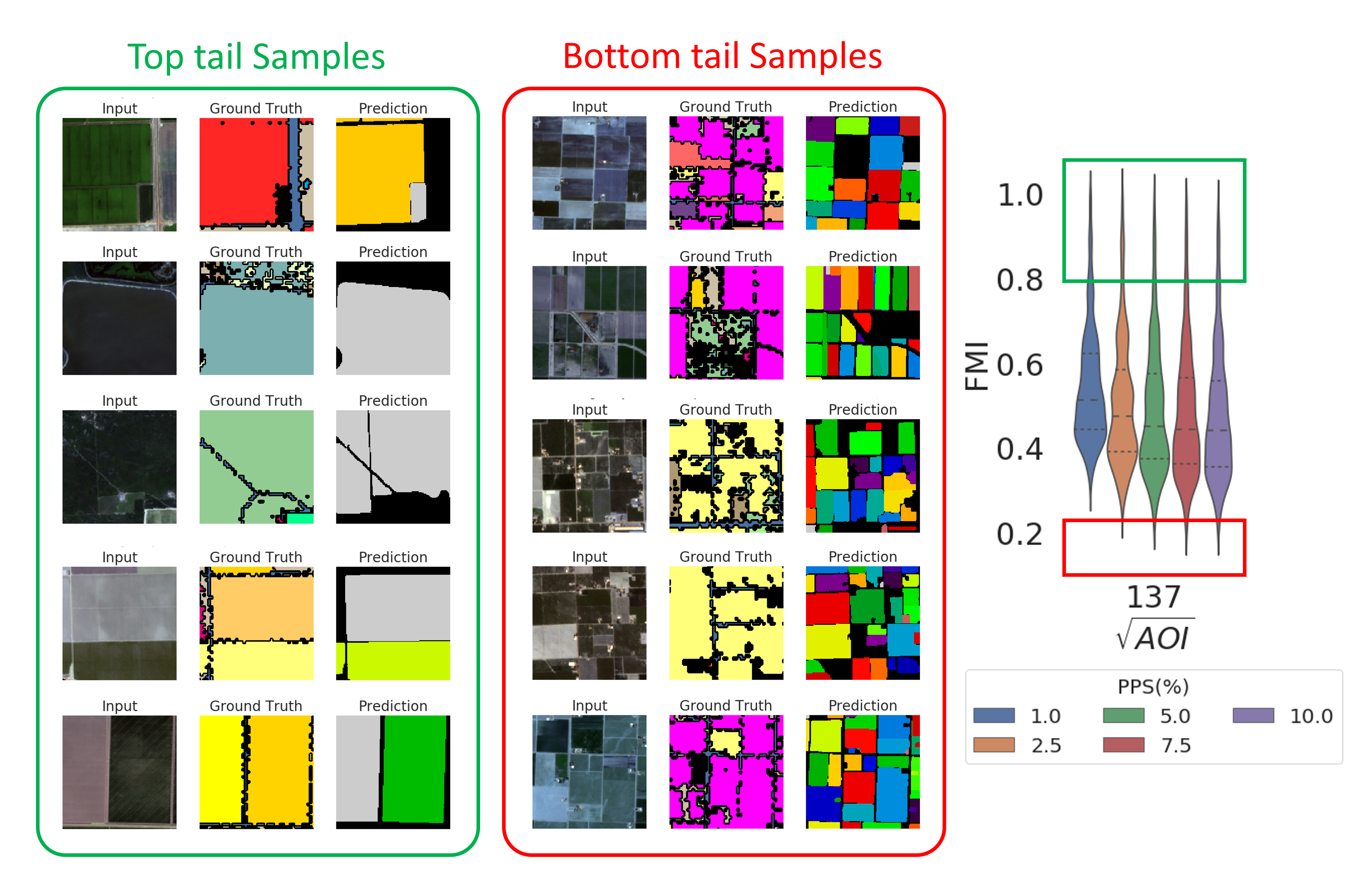}
  \caption{Examining samples from the tails of the distribution of FMI scores. \textbf{Top tails: (Leftmost plots, green outline)} Samples with high FMI scores tend to have semantically identical regions that remain spatially contiguous and occupy a large subarea in the AOI. \textbf{Bottom tails: (Middle plots, red outline)} Samples with low FMI scores tend to have semantically identical regions that don't remain spatially contiguous as they are separated by crop-field boundaries, roads, or other structures.}
  \label{fig:FMI_agg_with_tail_samples}
\end{figure}

Overall, \textbf{the average clustering consensus is low across all metrics} with really long tails which are indicative of outlier samples that align very well (or misalign terribly) with the ground-truth purely by chance across different AOIs (see figure \ref{fig:FMI_agg_with_tail_samples}). The highest scoring samples at the smallest AOI tend to have a few large semantically identical regions that remain spatially contiguous and therefore get segmented in a way that aligns with the ground-truth CDL (see figure \ref{fig:FMI_agg_with_tail_samples}, leftmost plots).

\begin{figure}[!htp]
  \centering
  \includegraphics[width=\textwidth]{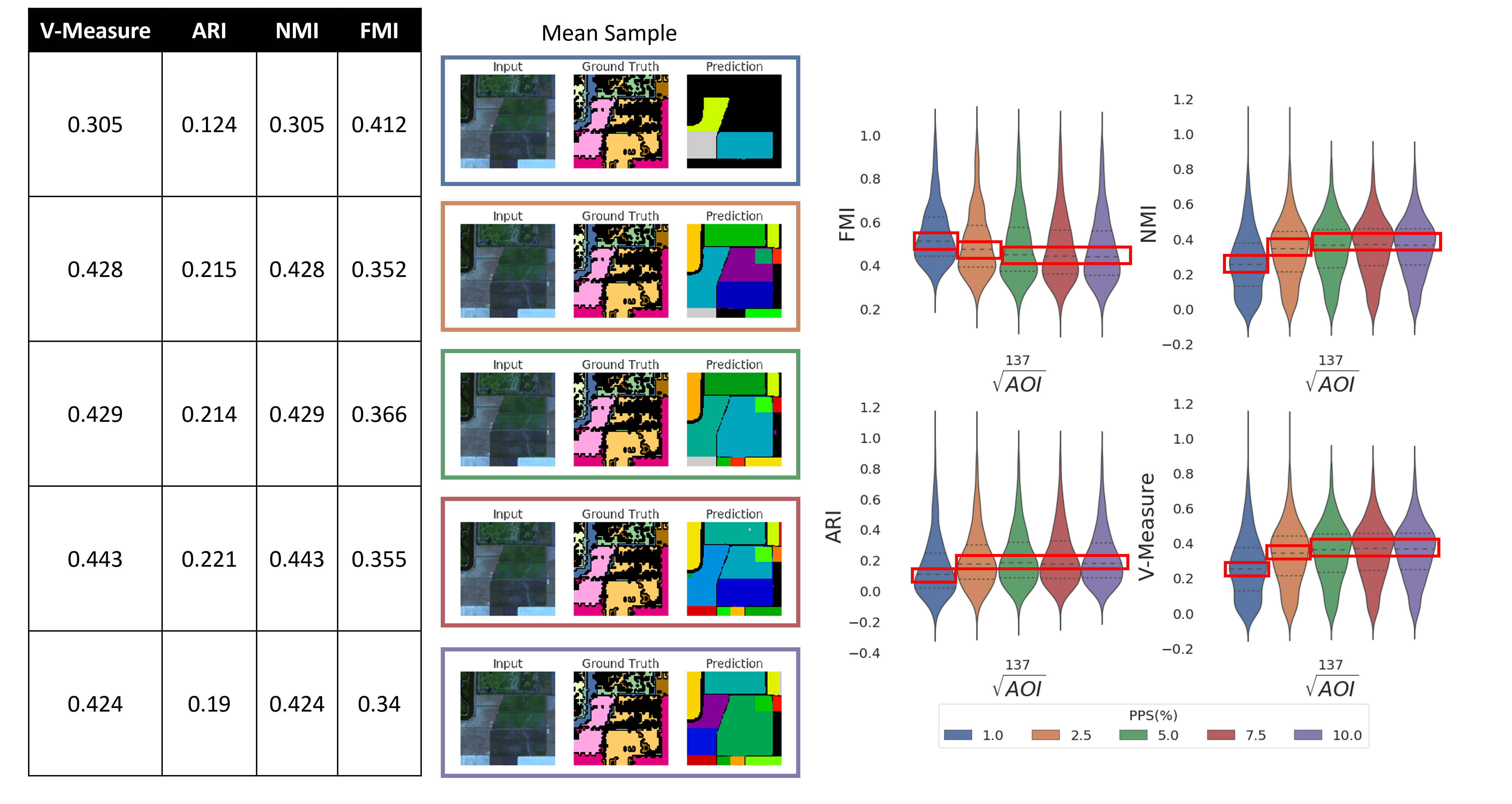}
  \caption{Examining a sample with mean FMI score over increasing prompts density (PPS\%): V-Measure, ARI and NMI improve as the prompt density increases while the FMI declines suggesting that although the clusters in the predicted mask become more homogenous and complete, the predicted mask captures the overall structure better in terms of pairwise agreements between cluster memberships of pixels and shared information between pixels in ground-truth and predicted masks, the predicted mask gets less accurate in terms of precision and recall.} 
  \label{fig:aggs_with_mean_samples}
\end{figure}

\subsection{SAM's Automatic Mask Generator for rapid, automatic crop fields shape-maps generation}
\begin{figure}[t!]
  \centering
  \includegraphics[width=0.9\textwidth]{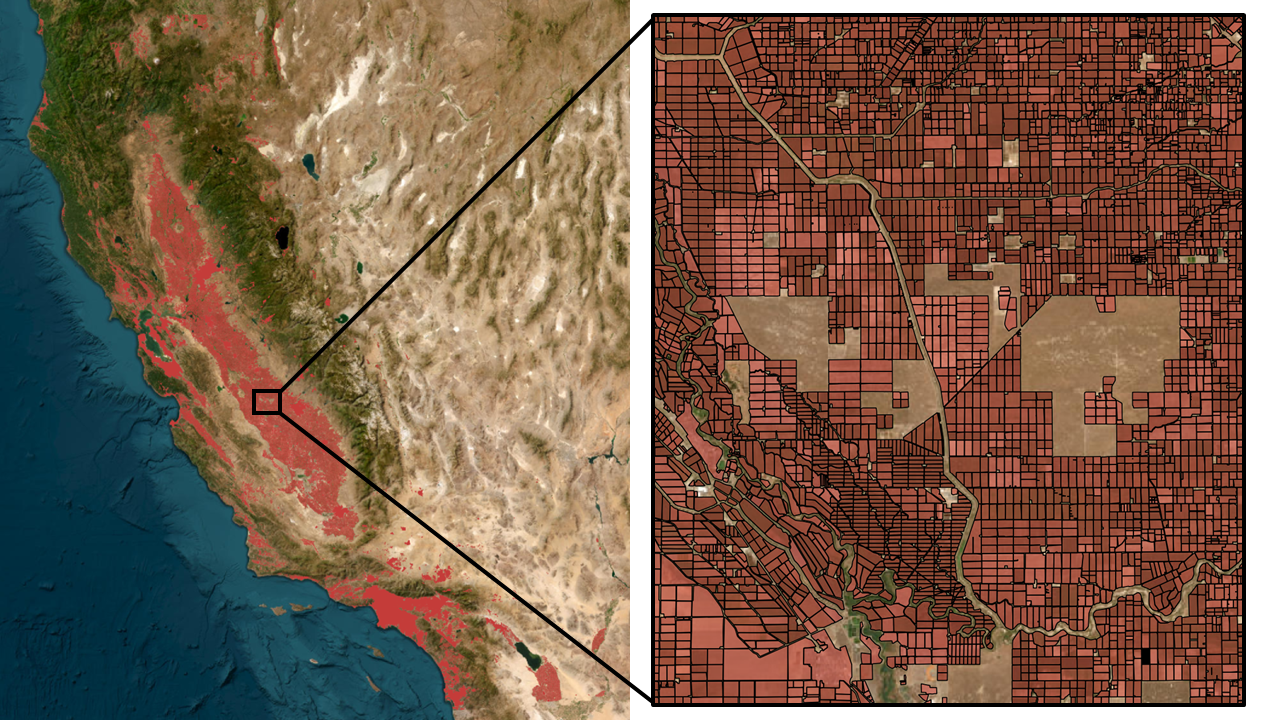} 
  \caption{An example AOI from agricultural fields of California overlaid with the shape-map - typically a shape file with a collection of polygons - depicting the extent and borders of crop fields in the AOI.}
  \label{fig:shape_map}
\end{figure}

The CDL is produced as a pixel-level classification by decision tree classifiers using the time-series and derived features of the temporal evolution of the NDVI information captured by LandSat satellite imagery \cite{boryan2011monitoring}. CDL is a high quality data product, however due to this pixel-level classification, which essentially lacks the spatial neighborhood context that modern deep convolutional neural networks use to great success for image processing tasks, the CDL has pixel noise (see figures \ref{fig:sample_pred} and \ref{fig:sample_pred_addtnl}, middle column). In modern efforts towards training a deep learning model to automatically predict the CDL using the satellite imagery input, a critical preprocessing step in producing good training examples is denoising the CDL of this pixel noise. There are various proposed methods for this preprocessing from manual approaches to more automatic supervised learning-based approaches \cite{ghosh2021calcrop21, zhang2020refinement, lin2022validation}. One convenient way to perform this preprocessing swiftly and accurately is to do field-level aggregation of the crop types, the reasoning being farmers typically plant a single crop in a field. To perform this field-level aggregation, shape maps of the crop fields in an AOI are used. For the state of California, these shape maps are produced by the California Department of Water Resources annually via on-site surveys as described \href{https://data.cnra.ca.gov/dataset/statewide-crop-mapping/resource/5cab9dde-5b20-4d2a-9e0c-993856e0898e}{here} (see figure \ref{fig:shape_map}). Based on analysis presented in this paper, we can envision a promising use-case for SAM to produce these shape maps of crop fields as the class-agnostic nature of zero-shot SAM's automatic mask generation applies more naturally to accelerating shape maps generation.
As shown in figure \ref{fig:sample_pred}(right column), \textbf{SAM is successful at identifying individual crop fields in an AOI separated by field borders, roads, waterways or other structures even though it struggles to identify semantically identical crop fields (i.e. fields containing the same crop-type) in an AOI as a single mask. Moreover, the resultant shape map is also low in pixel noise.} 

As CalCrop21 dataset does not provide the shape files corresponding to the the samples, we cannot perform a quantitative analysis for this proposed use-case and can only provide a qualitative assessment. Therefore, we leave the quantitative analysis to future work.

\section{Conclusions and Future Directions}
\label{sec:conc}
Our findings in this paper indicate that, while direct crop-type map generation using SAM's automatic mask generator (AMG) with uniformly distributed prompts is infeasible, we foresee a promising alternative in using it for shape maps generation instead. Our experiments demonstrate that SAM can be a valuable tool for producing fast and noise-free shape maps outlining individual fields within a large agricultural AOI in a satellite image. These shape maps, which are currently created manually as an annual data product, while not directly representing crop types can serve as a foundational step in the crop-type map generation process. Although, SAM's AMG enables swift annotations for features/objects of interest in the AOI, currently, the class-agnostic output limits us from predicting "true" multi-class masks where there is a one-to-one correspondence between the ground-truth and predicted labels. For a future direction, we can envision a use-case where we can use the ground-truth CDL to prompt SAM in a "CDL-informed" fashion one crop-type at a time and consolidate the class-wise binary masks into a "true" multi-class mask. In conclusion, our work provides steps towards bridging the gap between state-of-the-art image segmentation models like SAM and the specific needs of the agriculture industry, offering a potential avenue for more efficient and cost-effective tools for precision agriculture practices. Ultimately, our work contributes to the development of innovative solutions that enhance sustainability and productivity in farming while addressing the challenges of producing high-quality crop-type maps.

\bibliography{main}

\appendix

\section{Additional (Input, Ground Truth, SAM Prediction) examples}
\begin{figure}[H]
  \centering
  \includegraphics[width=\textwidth]{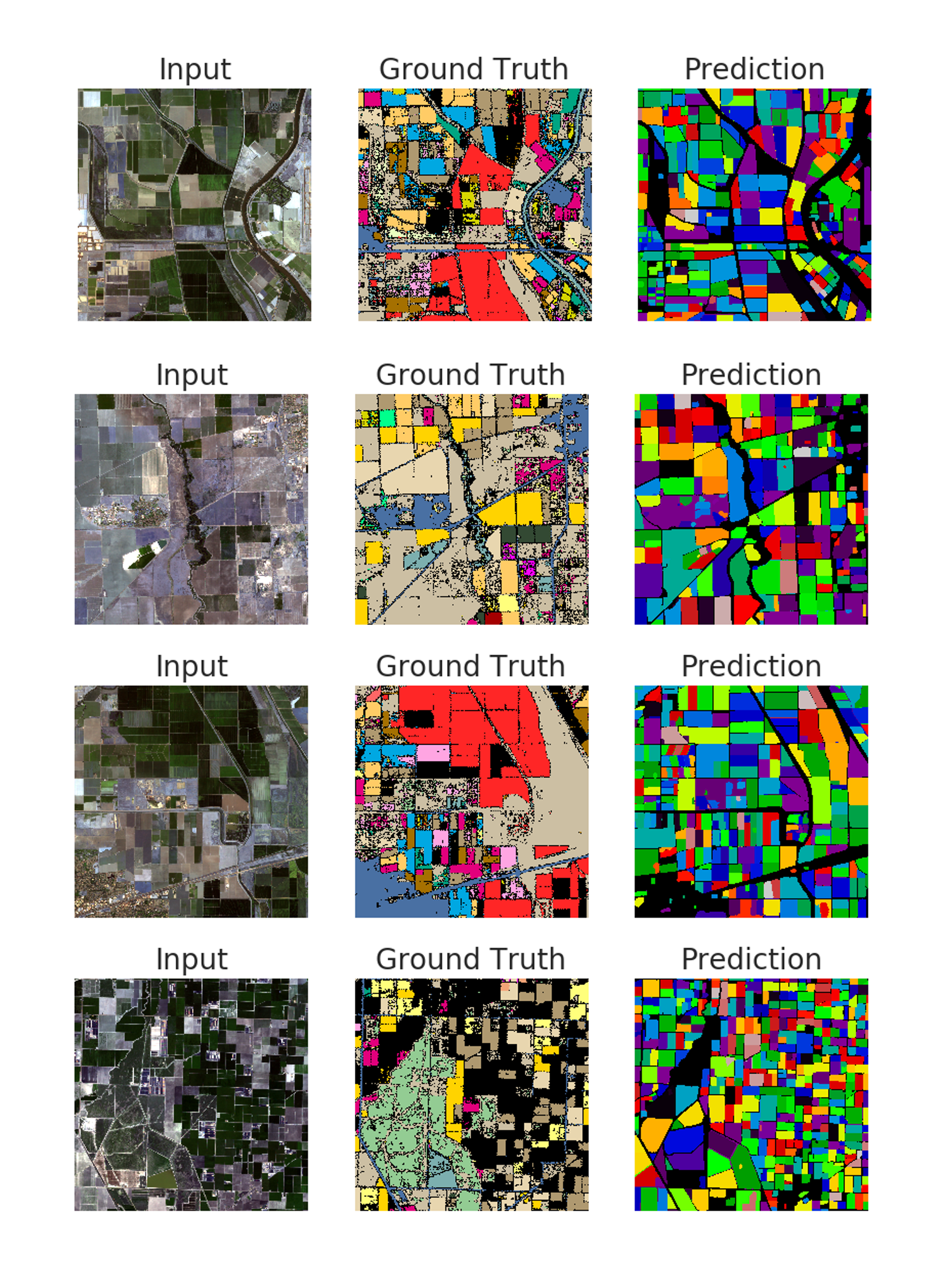}
  \caption{Additional examples: (left) Input images (AOI = 1098 x 1098) created using the red-green-blue channels at the maximum NDVI timestep from the 4D multispectral spatiotemporal imagery stack from Sentinel-2 satellite, (middle) the ground-truth crop-type maps (CDL) depicting crop types and other related classes, (right) zero-shot predicted masks using SAM's automatic mask generator.}
  \label{fig:sample_pred_addtnl}
\end{figure}

\section{Effect of varying MMRA on clustering consensus}
\label{ap:mmra_effect}
\begin{figure}[H]
  \centering
  \includegraphics[width=\textwidth]{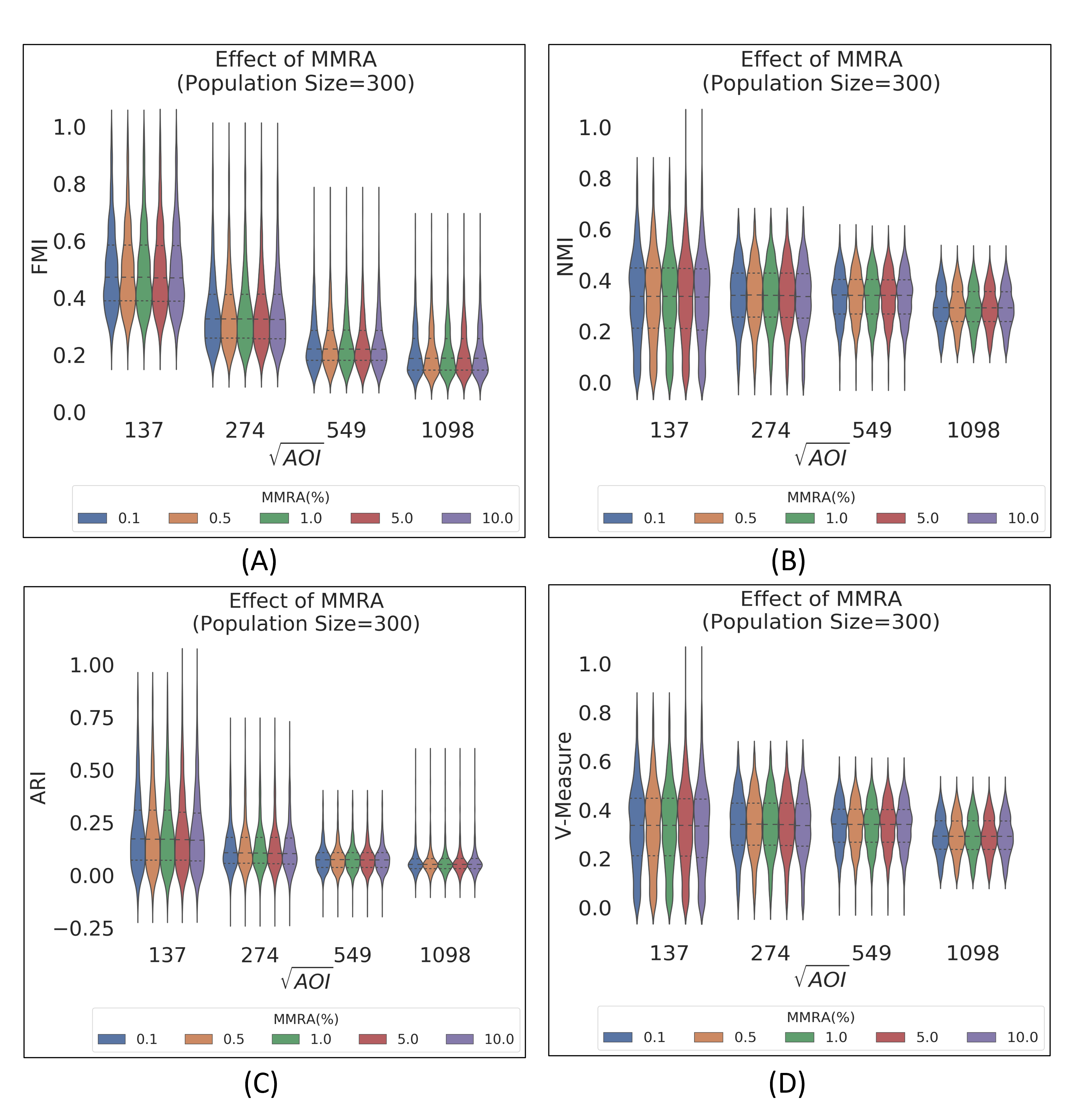} 
  \caption{MMRA does not have any effect on the mean scores for a given $\sqrt{AOI}$ across all 4 metrics.}
  \label{fig:perc_mmra_aggs}
\end{figure}

As shown in figure\ref{fig:perc_mmra_aggs}, the overall trend in FMI (and to a lesser extent ARI) shows decreasing consensus over increasing $\sqrt{AOI}$. However, for a given $\sqrt{AOI}$, unlike prompt density (PPS\%), varying the minimum mask region area (MMRA\%) to eliminate smaller masks did not demonstrate any effect across our experiments.

\section{Some high and low scoring examples}
\label{ap:examples_preds}
\begin{figure}[H]
  \hspace*{-1.5cm}
  \includegraphics[width=1.2\textwidth]{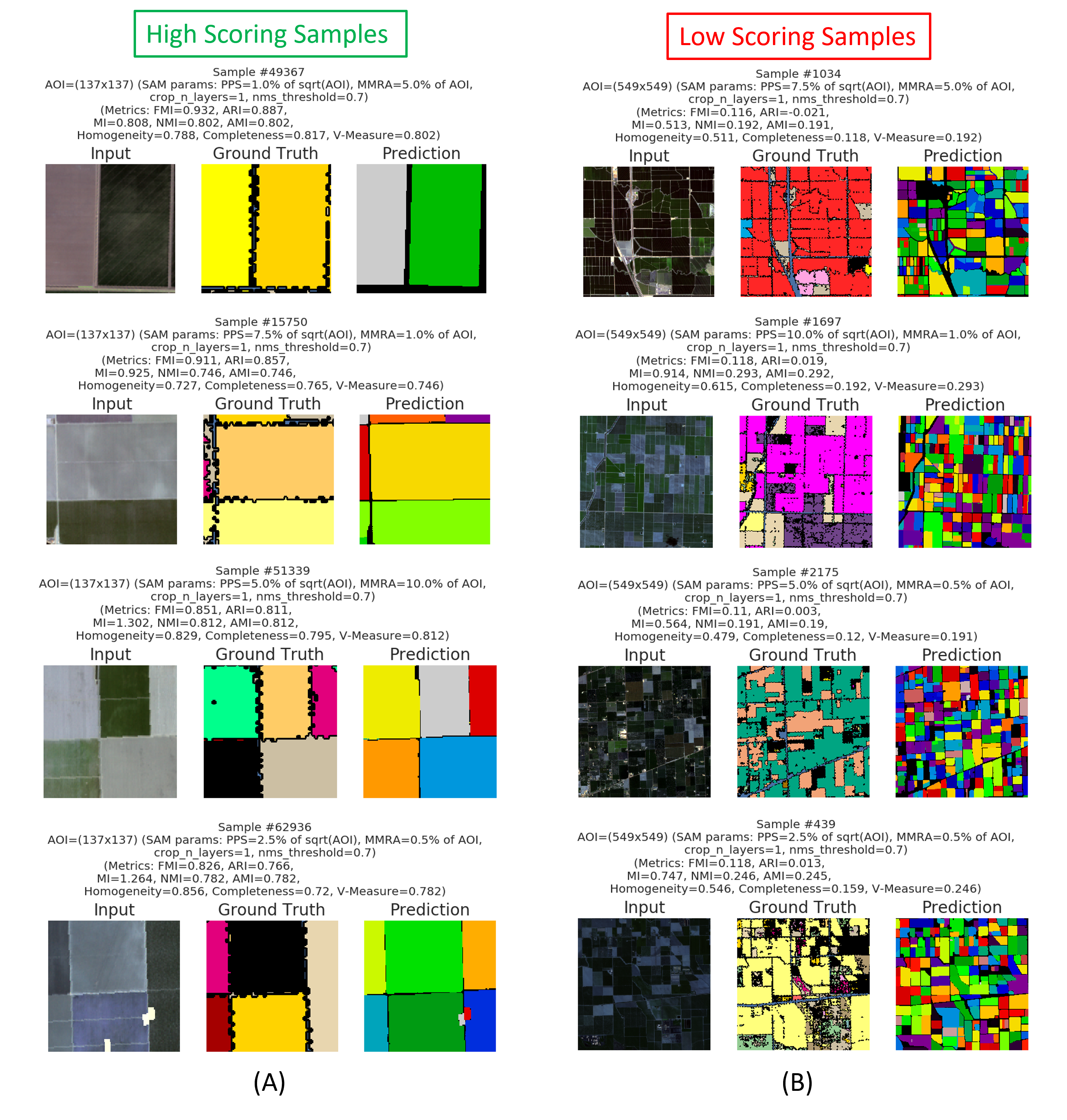} 
  \caption{Samples where semantically identical regions remain spatially contiguous and span a large subarea in the AOI are segmented appropriately by SAM in its zero-shot setting with uniformly distributed prompts.}
  \label{fig:easy_and_hard_samples}
\end{figure}

Figure \ref{fig:easy_and_hard_samples}(A) shows a selection of samples with high consensus scores while figure \ref{fig:easy_and_hard_samples}(B) shows a selection of low scoring samples.

\end{document}